\documentclass[
]{ceurart}

\usepackage{booktabs}
\usepackage{array}
\usepackage{makecell}
\usepackage{tabularx}
\usepackage{longtable}
\usepackage[table]{xcolor}
\usepackage{tcolorbox}
\usepackage{tikz}
\usepackage{pgfplots}
\usetikzlibrary{arrows.meta, positioning, shapes.geometric, calc}
\pgfplotsset{compat=1.18}
\newcolumntype{P}[1]{>{\raggedright\arraybackslash}p{#1}}
\definecolor{castonBlue}{HTML}{245B8F}
\definecolor{castonTeal}{HTML}{2A9D8F}
\definecolor{castonGold}{HTML}{D99A2B}
\definecolor{castonRed}{HTML}{B44A4A}
\definecolor{castonGray}{HTML}{5F6B73}
\definecolor{castonLight}{HTML}{EEF4F7}
\newtcolorbox{promptbox}[1][]{%
  colback=castonLight,
  colframe=castonBlue!70!black,
  boxrule=0.45pt,
  arc=2pt,
  left=5pt,
  right=5pt,
  top=5pt,
  bottom=5pt,
  fonttitle=\bfseries\sffamily,
  coltitle=black,
  #1
}

\begin{document}

\copyrightyear{2026}
\copyrightclause{Copyright for this paper by its authors. Use permitted under Creative Commons License Attribution 4.0 International (CC BY 4.0).}
\conference{ISWC 2026 Companion Volume, October 25--29, 2026, Bari, Italy}

\title{Ontology Refinement with Embeddings and LLMs: A Case Study on CASTON}

\author[1]{Hui Yang}[%
email=hui.yang-2@manchester.ac.uk,
]
\cormark[1]
\author[3,5]{Ziyu Li}[%
email=ziyu.li@ju.se,
]
\author[4]{Xuan Liu}[%
email=xuanliu@u.nus.edu,
]
\author[3]{Anders Jarfors}[%
email=anders.jarfors@ju.se,
]
\author[2]{He Tan}[%
email=tan.he@ju.se,
]
\author[1]{Jiaoyan Chen}[
email=jiaoyan.chen@manchester.ac.uk,
]
\cormark[1]

\cortext[1]{Corresponding author.}

\address[1]{University of Manchester, Manchester, UK}
\address[2]{Department of Computing, J\"onk\"oping University, Sweden}
\address[3]{Department of Materials and Manufacturing, J\"onk\"oping University, Sweden}
\address[4]{Department of Mechanical Engineering, National University of Singapore, Singapore}
\address[5]{COMPtech i Skillingaryd AB, Sweden}
\begin{abstract}
Ontologies are widely used to represent and organize complex knowledge in domains such as biomedicine and healthcare. However, constructing reliable ontologies remains challenging, as class hierarchies, whether manually or automatically constructed, may contain subtle subsumption errors that require careful refinement.
We study this problem in CASTON, a casting-domain ontology, and present a lightweight but scalable refinement workflow that combines a fine-tuned ontology embedding model with Large Language Models (LLMs). The embedding model is used to extract a limited number of candidate axioms for refinement together with possible replacement candidates, and an LLM then returns repair suggestions.
In a small expert evaluation, the default configuration using a CASTON-fine-tuned OnT embedding model and Qwen3-8B produced reliable high-confidence repairs: all six high-confidence replacements were accepted by both domain experts. The implementation is available at \url{https://github.com/HuiYang1997/OntRefine}.
\end{abstract}


\maketitle

\section{Introduction}

OWL ontologies provide a formal way to represent domain knowledge through concepts, relations, and logical constraints. They support knowledge sharing, automated reasoning, and explainable systems across various real-world domains, such as biomedicine~\cite{ashburner2000gene} and healthcare~\cite{SNOMED}.
However, building and maintaining ontologies still requires considerable expert effort. In particular, incorrect asserted superclass axioms can affect reasoning results and downstream applications, making class hierarchy refinement an important but time-consuming task.

Recent studies have explored the use of Large Language Models (LLMs) for ontology completion, matching, and validation~\cite{chen2021ontolama,he2023language,ont2025,llms4ol2023}. However, asking an LLM to inspect a whole ontology in order to refine subsumption axioms is not well targeted and can be computationally expensive. Therefore, we propose a two-stage approach.
In the first stage, an embedding model helps narrow down the search by ranking the existing axioms and retrieving possible alternative parent classes.
In the second stage, an LLM checks only these selected axiom groups. It compares the current parent class with the candidate parent classes using their class labels, and then gives structured refinement suggestions.
This design keeps the LLM focused on a small local context. It also enhances the traceability of each suggested repair, because every repair is linked to a clear and limited set of candidate classes.

We use CASTON\footnote{\url{https://inmost-research.org/ontologies/caston/}}, a developing real-world ontology for metal casting, materials science, and manufacturing processes, as a case study to evaluate whether embedding-guided LLM review can support the refinement of asserted superclass axioms. Using the CASTON 1.0.0 snapshot released on 22 February 2026, %
we evaluated different combinations of embedding models and LLMs.
Across these combinations, the system produced 40 replacement suggestions with high or medium confidence. Human evaluation by two domain experts shows that the default setting, which combines a CASTON-fine-tuned embedding model with Qwen3-8B, provides reliable high-confidence repairs: all six evaluated high-confidence replacements were unanimously accepted. These findings indicate that embedding-guided LLM review can provide practical support for ontology construction by prioritising likely problematic superclass axioms and suggesting localised repairs, thereby reducing manual inspection effort and improving scalability in large ontology development settings.

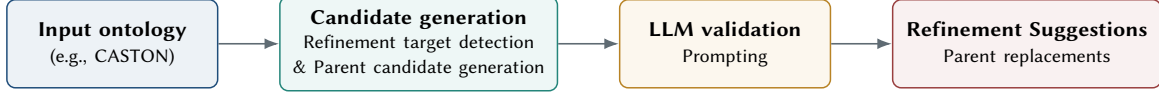
\begin{figure}[t]
\centering
\resizebox{0.96\linewidth}{!}{%
\begin{tikzpicture}[
    font=\sffamily\small,
    node distance=10mm,
    stage/.style={
        rounded corners=4pt,
        draw=#1!85!black,
        fill=#1!8,
        line width=0.7pt,
        minimum height=1.45cm,
        text width=3.05cm,
        align=center,
        inner sep=6pt
    },
    flow/.style={
        -{Latex[length=2.4mm,width=1.8mm]},
        line width=0.75pt,
        draw=castonGray
    }
]

\node[stage=castonBlue] (input)
{\textbf{Input ontology}\\
\footnotesize (e.g., CASTON)};

\node[stage=castonTeal, right=of input, text width=4.15cm] (candidate)
{\textbf{Candidate generation}\\
\footnotesize Refinement target detection\\
\footnotesize \& Parent candidate generation};

\node[stage=castonGold, right=of candidate] (validation)
{\textbf{LLM validation}\\
\footnotesize Prompting};

\node[stage=castonRed, right=of validation, text width=4cm] (output)
{\textbf{Refinement Suggestions}\\
\footnotesize Parent replacements};

\draw[flow] (input) -- (candidate);
\draw[flow] (candidate) -- (validation);
\draw[flow] (validation) -- (output);

\end{tikzpicture}%
}
\caption{Overview of the ontology-refinement workflow.} %
\label{fig:workflow}
\end{figure}

\section{Method}
We focus on asserted in-ontology subsumption axioms between named CASTON classes.
That is, each axiom has the form $A\sqsubseteq B$, where $A$ and $B$ are
named CASTON classes and the axiom states that every instance of $A$ is
also an instance of $B$. As shown in Figure~\ref{fig:workflow}, the
workflow has two stages: embedding-based candidate generation and LLM
validation.

\paragraph{Stage 1: Candidate generation.}
For each asserted axiom $A \sqsubseteq B$, %
we rank all named classes $B' \in \mathcal{O}$, excluding $A$, as possible parents of $A$.
Let $r_m(A,B)$ denote the rank assigned by embedding model $m$ to the asserted parent $B$ of class $A$, with rank 1 indicating the most plausible parent. Formally,
$$
r_m(A,B)=\operatorname{rank}_{s_m(A,B)}\{s_m(A,B')| \textit{ all classes } B' \neq A\},
$$
where $s_m(A,B')$ is the score function of the embedding model $m$ for measuring the plausibility of the axiom $A \sqsubseteq B'$.
Thus, a larger value of $r_m(A,B)$ indicates that the asserted parent is less compatible with the embedding model.
We then select the top \(k_r\) axioms with the largest \(r_m(A,B)\) values, using \(k_r=30\) by default, as suspicious axioms for LLM-based validation.

For each selected axiom $A \sqsubseteq B$, we use the same embedding model to construct a pool of alternative candidate parents. Specifically, for the child class $A$, we retain the $k_a$ (with $k_a = 10$ by default) highest-ranked candidate parents that are neither $A$ itself nor already asserted parents of $A$. We denote this set by $P_{\mathrm{cand}}(A)$.

As the default embedding model, we use OnT~\cite{ont2025} after fine-tuning it on CASTON. OnT is a hyperbolic ontology embedding method designed to adapt to a given \(\mathcal{EL}\) ontology. It models a score function $s(A, B)$ based on the hyperbolic distance between the embeddings of \(A\) and \(B\).
Since CASTON falls within this ontology class, OnT is suitable for modelling its hierarchical structure.

\paragraph{Stage 2: LLM validation.}
For each axiom $A \sqsubseteq B$ selected in the previous stage, we construct a single LLM prompt that combines three sources of information: the child class $A$, its currently asserted superclass axioms, and the candidate parent set $P_{\mathrm{cand}}(A)$, as illustrated in Figure~\ref{fig:prompt}. For each class mentioned in the prompt, we include the available metadata, such as the class ID, label, definition or comment, and domain or scope note. The prompt also includes all other additional axioms involving $A$, such as existential restrictions of the form $A \sqsubseteq \exists r. B_1$, as background information where available. %

\begin{figure}[htb]
    \centering

    \begin{promptbox}[title={User Prompt Template}]
    \footnotesize

    \textbf{Concept under review.}
    The prompt provides the ID, label, definition, domain, and other axioms
    for the child class $A$.\\[0.3pt]
    \textbf{Current superclass axioms.} Lists the axioms
    $A \sqsubseteq B_1,\ldots, A\sqsubseteq B_n$, together with the label/definition of $B_i$.\\[0.3pt]
    \textbf{Top candidate parent classes.}
    The top $k=10$ candidate parents
    $B'$ that are not in current parents.\\[0.3pt]
    \textbf{Decision rule.}
    A direct parent must satisfy the following condition: every instance of
    $A$ is necessarily an instance of $B$. Multiple inheritance is allowed.\\[0.3pt]
    \textbf{Required JSON schema.}
{\ttfamily\scriptsize \begin{quote} \{\\ \hspace*{1em}"concept": "\textless class\_id\textgreater",\\ \hspace*{1em}"verdict\_per\_parent": [\\ \hspace*{2em}\{"parent": "\textless parent\_id\textgreater", "verdict": "correct|incorrect|partial",\\ \hspace*{2.4em}"confidence": "high|medium|low", "reason": "\textless brief explanation\textgreater"\}\\ \hspace*{1em}],\\ \hspace*{1em}"suggested\_replacements": [\\ \hspace*{2em}\{"remove": "\textless parent\_id\textgreater", "add": "\textless new\_parent\_id\textgreater", "reason": "\textless explanation\textgreater"\}\\ \hspace*{1em}],  \\
 \} \end{quote} }

    \end{promptbox}

    \caption{Prompt used for the LLM validation stage.}
    \label{fig:prompt}
\end{figure}

The LLM is asked to determine whether each current parent is a correct direct superclass, whether any parent should be removed or replaced by one of the suggested alternatives, whether any candidate parent should be added, and to provide a justification for each decision. The model returns a structured JSON object containing the parent-level verdict, a suggested replacement where applicable, a textual justification, and a confidence label: \textsc{High}, \textsc{Medium}, or \textsc{Low}. We retain suggestions with \textsc{High} or \textsc{Medium} confidence as candidate refinements for subsequent evaluation.

\section{Experimental Setup and Preliminary Results}
\label{sec:evaluation}
\paragraph{Experiment setting.}
\textbf{(1) Dataset:}
We evaluate our method on the developing ontology CASTON, using the first release, version 1.0.0, which was released on 22 February 2026. This version contains 270 named classes, i.e., concepts, and 306 axioms. Of these, 292 are subsumption axioms involving atomic concepts, for which rank scores were assigned using the embedding models, as illustrated in the workflow.
\textbf{(2) Configuration.}
As shown in the first column of Table \ref{tab:source-confidence-summary}. We consider three different settings:
\textit{The default setting} is CASTON-fine-tuned OnT + Qwen3-8B, where OnT is fine-tuned from the SBERT variant all-mpnet-base-v2. To analyze the influence of embedding models and LLMs, we compare the default workflow with \textit{two non-fine-tuned comparison settings using Qwen3-8B or GPT-4o mini}, where both method use
 a non-CASTON-fine-tuned ensemble embedding for candidate generation. The ensemble method composed of three sentence embedding models: OnT fine-tuned on the GALEN ontology, HiT fine-tuned on WordNetNoun, and standard SBERT. All three models are based on all-mpnet-base-v2. In this ensemble setting, we use the mean rank across the embedding models to select candidate refinement axioms and generate candidate parent sets. %
\textbf{(3) Evaluation metric.}
Two domain experts independently score each replacement suggestion using a five-point ordinal scale: \(-2\) for completely wrong, \(-1\) for seems wrong, \(0\) for vague or unclear, \(+1\) for seems correct, and \(+2\) for completely correct. We classify a suggestion as accepted when both experts assign positive scores, rejected when both assign negative scores, and mixed otherwise.

\paragraph{Results}

Table~\ref{tab:source-confidence-summary} shows that the default configuration is highly reliable for the small set of high-confidence replacement suggestions evaluated in this study.
With Qwen3-8B and fine-tuned OnT embeddings, all six high-confidence replacements are accepted by both experts (with score +2 for all cases), yielding a 100\% acceptance rate with no mixed or rejected cases. In contrast, both medium-confidence cases are mixed. This indicates that the LLM confidence score is useful for prioritization: high-confidence predictions under the default setting are trustworthy, whereas medium-confidence ones still require manual review.

Replacing the fine-tuned embeddings with the pretrained ensemble markedly reduces reliability. With the same Qwen3-8B generator, the high-confidence acceptance rate drops from 100\% to 57.1\%: 4 of 7 cases are accepted, while 3 are rejected. Medium-confidence results are also weak, with only 1 of 4 cases accepted and 3 rejected. Thus, although the pretrained ensemble can recover some valid candidates, it lacks the precision of the fine-tuned OnT model, highlighting the importance of task-specific embedding finetuning.

Using GPT-4o mini with the same pretrained ensemble further reduces the reliability of high-confidence predictions. Among the 16 high-confidence replacements, only 5 are consistently accepted, 1 receives mixed judgments, and 10 are rejected, yielding an acceptance rate of 31.3\%. This is 68.7 percentage points lower than the default setting. Notably, many replacements assigned high confidence by GPT-4o mini are rejected by both experts.
These results indicate that, compared with Qwen3-8B, GPT-4o mini exhibits poorer confidence calibration for this task and a stronger tendency to produce unreliable decisions.

\begin{table}[t]
\centering
\caption{Expert outcomes by setting and LLM confidence. Accepted rows have positive expert agreement; parentheses give the +2/+1 breakdown.
Mixed rows are expert disagreements or unclear cases.}
\label{tab:source-confidence-summary}
\scriptsize
\setlength{\tabcolsep}{3.5pt}
\renewcommand{\arraystretch}{1.08}
\begin{tabular}{@{}llrcccc@{}}
\toprule
\textbf{Setting} & \textbf{Confidence} & \textbf{Rows} &
\makecell{\textbf{Accepted}\\\textbf{(+2/+1)}} &
\textbf{Mixed} & \makecell{\textbf{Rejected}\\\textbf{(-2/-1)}} & \textbf{Acceptance rate} \\
\midrule
\rowcolor{castonTeal!12}
Default: Qwen3-8B + CASTON OnT-FT & High & 6 &
\cellcolor{castonTeal!70}\textbf{6} (6/0) & 0 & 0 & \textbf{100\%} \\
\rowcolor{castonTeal!5}
Default: Qwen3-8B + CASTON OnT-FT & Medium & 2 &
0 (0/0) & \cellcolor{castonGray!32}\textbf{2} & 0 & 0\% \\
\addlinespace[1pt]
Ablation: Qwen3-8B + pretrained ensemble & High & 7 &
\cellcolor{castonTeal!42}\textbf{4} (4/0) & 0 &
\cellcolor{castonRed!36}\textbf{3} & 57.1\% \\
Ablation: Qwen3-8B + pretrained ensemble & Medium & 4 &
\cellcolor{castonTeal!24}1 (0/1) & 0 &
\cellcolor{castonRed!36}\textbf{3} & 25.0\% \\
\addlinespace[1pt]
Auxiliary: GPT-4o mini + pretrained ensemble & High & 16 &
\cellcolor{castonTeal!42}\textbf{5} (5/0) &
\cellcolor{castonGray!18}1 &
\cellcolor{castonRed!50}\textbf{10} & 31.3\% \\
Auxiliary: GPT-4o mini + pretrained ensemble & Medium & 5 &
\cellcolor{castonTeal!24}1 (0/1) & 0 &
\cellcolor{castonRed!40}\textbf{4} & 20.0\% \\
\bottomrule
\end{tabular}
\end{table}

\paragraph{Case Studies}
\label{sec:evaluation-examples}
To better understand the strengths and limitations of the proposed workflow, we present detailed case studies generated using the default configuration (i.e., Qwen3-8B+OnT-FT).

\textit{Success case:} A representative successful refinement is the concept \texttt{chromite\_sand}. In the original ontology, \texttt{chromite\_sand} was modelled as a subclass of \texttt{green\_sand\_molding}. The fine-tuned workflow instead suggested replacing its superclass with \texttt{foundry\_sand}, a recommendation accepted by both domain experts (+2/+2). Semantically, \texttt{green\_sand\_molding} denotes a molding process, whereas \texttt{chromite\_sand} is a mold material used in foundry applications. The suggested axiom therefore yields a more accurate material taxonomy while preserving the intended domain semantics. Similar refinements were observed for \texttt{olivine\_sand}, \texttt{silica\_sand}, and \texttt{zircon\_sand}, indicating that the fine-tuned OnT embeddings can identify semantically appropriate superclass refinements rather than merely retrieving lexically related concepts.

\textit{Failure case:} A representative failure case is \texttt{semisolid\_metal\_(ssm)\_processing}. The original ontology classified this concept under \texttt{high\_pressure\_die\_casting}, whereas the fine-tuned workflow suggested replacing its superclass with \texttt{semisolid\_casting}. Although this suggestion is semantically related, the experts did not agree on its validity (0/-2). One expert noted that the hierarchy among these concepts is inconsistent in the current ontology and requires revision, while the other argued that \texttt{semisolid\_metal\_processing} should instead be modelled as a subclass of \texttt{die\_casting}. This case shows that even fine-tuned embeddings may produce incorrect superclass assignments when the existing ontology hierarchy is incomplete or structurally inconsistent.

\section{Conclusion}

This poster provides preliminary evidence that embedding-guided LLM review can support practical ontology hierarchy refinement when the embedding model is adapted to the target ontology.
The CASTON-fine-tuned OnT + Qwen3-8B workflow identified a small but reliable set of high-confidence repairs, all of which were judged to be completely correct by two domain experts. The evaluation is currently limited to one ontology, replacement suggestions, and a small number of expert-assessed repairs.
In future work, we plan to extend this pipeline to more complex axiom types, such as relational restrictions, and evaluate its effectiveness at larger scale across ontologies from diverse domains. We will also apply the pipeline to future versions of CASTON to support systematic hierarchy refinement, ontology enrichment, and continuous ontology evolution through expert-guided, LLM-assisted refinement.

\section*{Acknowledgments}

Hui Yang and Jiaoyan Chen are supported by the EPSRC project OntoEm (EP/Y017706/1). Xuan Liu acknowledges the support of the NUS Research Scholarship. Ziyu Li's PhD project was funded by the Stiftelsen för kunskaps- och kompetensutveckling (KK-Stiftelsen) through the Smart Industry Sweden initiative (project number 2020-0044).

\section*{Declaration on Generative AI}

During the preparation of this work, the authors used ChatGPT for \textit{Grammar and spelling check}, \textit{Paraphrase and reword}, and \textit{Improve writing style} to check grammar and spelling and improve the clarity and readability of author-written text. Claude was used to assist in developing code implementations for the proposed methods. After using these tools and services, the authors reviewed and edited the generated content as needed and take full responsibility for the publication's content.

\bibliography{references}
\end{document}